\documentclass[singlecolumn,10pt]{article}
\usepackage{geometry}

\usepackage[table,dvipsnames]{xcolor}
\usepackage{colortbl}
\definecolor{mygray}{gray}{0.93}

\usepackage[numbers]{natbib}
\usepackage{microtype}
\usepackage{graphicx}
\usepackage[inline]{enumitem}
\usepackage{subfig}
\usepackage{booktabs} 
\usepackage{array}
\usepackage{multirow}
\usepackage{bm}
\usepackage{tabularx}

\usepackage{tikz}
\usepackage{amsmath}
\usepackage{hyperref} 

\begin{document}

\title{
    \Large \bf
    \rule{\textwidth}{2pt} \\[10pt]
    ChocoLlama: Lessons Learned From Teaching Llamas Dutch \\
    \rule{\textwidth}{2pt}
}

\date{}

\author{
  Matthieu Meeus\textsuperscript{1}\thanks{Equal contribution.} \and
  Anthony Rathé\textsuperscript{2}\footnotemark[1] \and
  François Remy\textsuperscript{3} \and 
  Pieter Delobelle\textsuperscript{4}\thanks{Work done while at KU Leuven.} \and
  Jens-Joris Decorte\textsuperscript{3,5,6} \hspace{0.5cm}
  Thomas Demeester\textsuperscript{3,5} \\ [10pt]
  \textsuperscript{1}\textit{Imperial College London} \hspace{1cm}
  \textsuperscript{2}\textit{Cavell} \hspace{1cm}
  \textsuperscript{3}\textit{Ghent University} \\[10pt]
  \textsuperscript{4}\textit{Aleph Alpha} \hspace{1cm}
  \textsuperscript{5}\textit{imec} \hspace{1cm}
  \textsuperscript{6}\textit{TechWolf}
}

\maketitle


\begin{abstract}

While Large Language Models (LLMs) have shown remarkable capabilities in natural language understanding and generation, their performance often lags in lower-resource, non-English languages due to biases in the training data. In this work, we explore strategies for adapting the primarily English LLMs (Llama-2 and Llama-3) to Dutch, a language spoken by 30 million people worldwide yet often underrepresented in LLM development. We collect 104GB of Dutch text ($32$B tokens) from various sources to first apply continued pretraining using low-rank adaptation (LoRA), complemented with Dutch posttraining strategies provided by prior work. For Llama-2, we consider using 
\begin{enumerate*}[label=(\roman*)]
\item the tokenizer of the original model, and 
\item training a new, Dutch-specific tokenizer combined with embedding reinitialization.
\end{enumerate*}
We evaluate our adapted models, ChocoLlama-2, both on standard benchmarks and a novel Dutch benchmark, ChocoLlama-Bench. Our results demonstrate that LoRA can effectively scale for language adaptation, and that tokenizer modification with careful weight reinitialization can improve performance. Notably, Llama-3 was released during the course of this project and, upon evaluation, demonstrated superior Dutch capabilities compared to our Dutch-adapted versions of Llama-2. We hence apply the same adaptation technique to Llama-3, using its original tokenizer. While our adaptation methods enhanced Llama-2's Dutch capabilities, we found limited gains when applying the same techniques to Llama-3. This suggests that for ever improving, multilingual foundation models, language adaptation techniques may benefit more from focusing on language-specific posttraining rather than on continued pretraining. We hope this work contributes to the broader understanding of adapting LLMs to lower-resource languages, and to the development of Dutch LLMs in particular. We provide open-source access to our models, benchmarks\footnote{\url{https://huggingface.co/ChocoLlama}} and code\footnote{\url{https://github.com/ChocoLlamaModel/ChocoLlama}}.

\end{abstract}

\section{Introduction}

The research and development of ever improving Large Language Models (LLMs) has expanded rapidly throughout the last years. Models such as GPT-3/4~\cite{brown2020language,gpt4techreport}, BLOOM~\cite{scao2022bloom}, Mistral 7B~\cite{jiang2023mistral}, Llama-1/2/3~\cite{touvron2023llama,touvron2023llama2,llama3modelcard} promise to revolutionise how machines are able to understand and generate natural language. 

LLMs acquire their capabilities from immense sets of training data, the vast majority of which has been scraped from the web. For instance, the frequently updated web-scraped dataset Common Crawl~\cite{commoncrawl} represents the majority of the training data for GPT-3~\cite{brown2020language}, Llama~\cite{touvron2023llama} and BLOOM~\cite{scao2022bloom}. Textual data on the web, however, represents a highly skewed language distribution. OSCAR~\cite{suarez2019asynchronous} classifies Common Crawl~\cite{commoncrawl} by language and finds that more than one third of all data represents English, with the remaining textual data distributed over more than $170$ other natural languages. This inherent bias in available textual data is propagated to the often limited multilingual capabilities of the best LLMs~\cite{bang2023multitask}.

\begin{figure}[t]
\centering
\includegraphics[width=\linewidth]{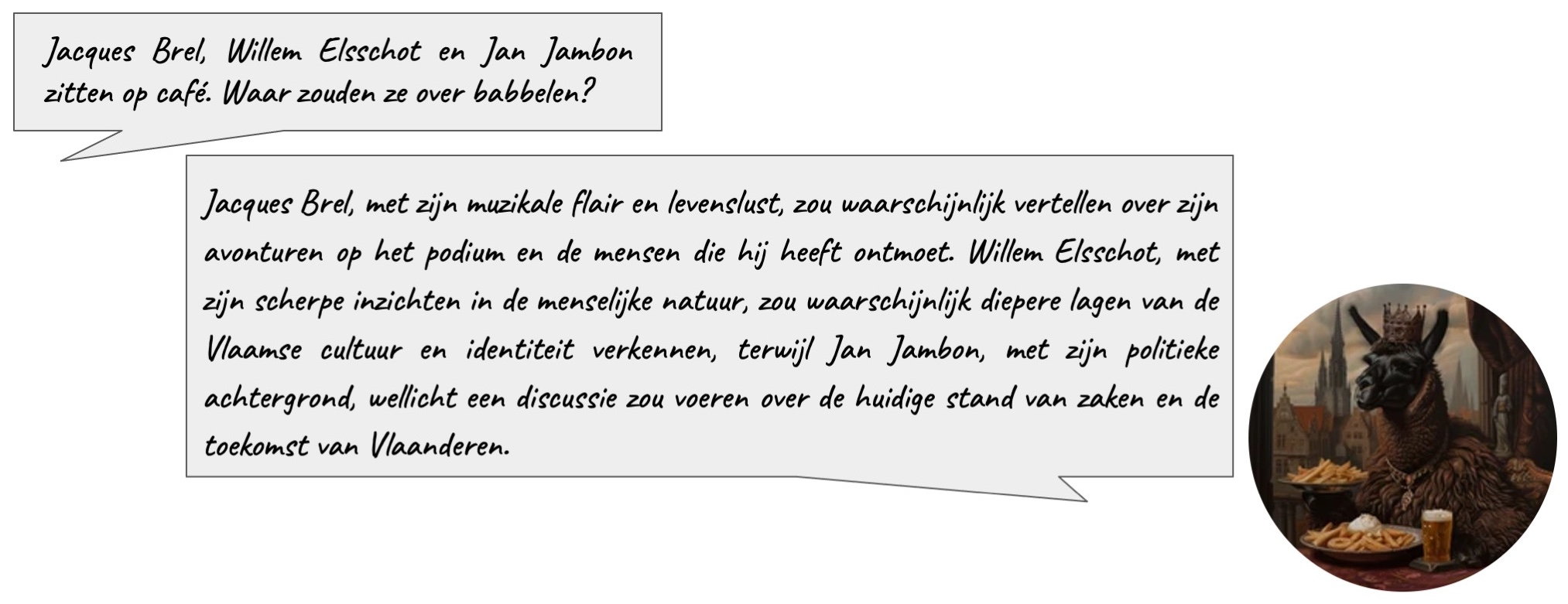} 
    \caption{Hand-picked conversational snippet of \textsf{Llama-3-ChocoLlama-instruct}. Answers to the same prompt for other models considered in this paper can be found in Appendix \ref{app:samples}.}
\label{fig:chat}
\end{figure}

In response to these limitations, model developers have either actively trained large-scale multilingual models~\cite{scao2022bloom,faysse2024croissantllm}, upsampled non-English data in the pretraining data corpus~\cite{jiang2024mixtral} or fine-tuned (primarily English) models to lower-resource languages~\cite{yong2022adapting,ebrahimi2021adapt}.

We here explicitly focus on Dutch, a natural language spoken by about 30 million people worldwide\footnote{\url{https://en.wikipedia.org/wiki/Dutch_language}}. While scarce, the research on Dutch language models has been expanding in the last years. Prior work has, for instance, trained masked language models from scratch using exclusively Dutch data~\cite{delobelle2020robbert,de2019bertje,devlin2018bert}. More recently, fine-tuned versions of state-of-the-art, primarily English LLMs have been made available on Hugging Face, including for instance GEITje-7B~\cite{rijgersberg2023geitje}, resulting from a continued pretraining of Mistral 7B~\cite{jiang2023mistral} on 10 billion tokens of Dutch data. GEITje has further been finetuned for conversational capabilities~\cite{geitjeultra}, and other models have been developed using similar finetuning methods and datasets without Dutch pretraining~\cite{reynaerdechat,llama3bdutch}.

\textbf{Contribution.} In this project, we explore different strategies to adapt the open-source, primarily English LLMs Llama-2~\cite{touvron2023llama2} and Llama-3~\cite{llama3modelcard}, developed by Meta AI, to Dutch.  

First, we collect $104$ GB of Dutch textual data ($32$B Llama-2~\cite{touvron2023llama2} tokens), most of which originating from the Dutch part of OSCAR~\cite{suarez2019asynchronous}, complemented with data from movie subtitles~\cite{lison2016opensubtitles2016} and books made available in the public domain~\cite{projectgutenberg}. We also collected three additional, publicly available Dutch datasets thanks to our industry partners. TechWolf\footnote{\url{https://techwolf.com/}} provided us with job descriptions collected over the last five years, while Bizzy\footnote{\url{https://bizzy.org/en}} provided legal documents related to Belgium-based corporations from ('\emph{Het Belgisch Staatblad}') and ML6\footnote{\url{https://www.ml6.eu/}} provided documents from Flemish legislation. 

Next, we use the parameter efficient finetuning method LoRA~\cite{hulora} to further pretrain Llama-2 7B~\cite{touvron2023llama2} for one epoch on our entire Dutch dataset. We first use the original tokenizer and model, resulting in \textsf{ChocoLlama-2-7B-base}. 

As the original model, and its tokenizer, have been primarily trained on English data, we also consider training a new BPE tokenizer from scratch and exclusively on Dutch data. To mitigate any forgetting associated with changing something as fundamental as the tokenizer (and vocabulary), we apply the embedding reinitialization approach as introduced by Remy et al.~\cite{remy2023tik}. We find the new tokenizer to compress the same amount of Dutch text into 29.4\% less tokens, and further pretrain the re-initialized Llama-2 again using LoRA to achieve \textsf{ChocoLlama-2-7B-tokentrans-base}. 

We further apply posttraining (supervised finetuning followed by Direct Preference Optimization~\cite{rafailov2024direct} updating all model parameters) as proposed by Vanroy~\cite{vanroy2023language}. The resulting models are available as \textsf{ChocoLlama-2-7B-instruct} and \textsf{ChocoLlama-2-7B-tokentrans-instruct}.

We evaluate our models across two benchmarks. First, we quantitatively evaluate the model performance on readily available Dutch translations of industry-standard benchmarks~\cite{lai2023openllmbenchmark,arc_benchmark,hellaswag_benchmark,mmlu_benchmark,truthfulqa_benchmark}. Second, we qualitatively evaluate models on a new benchmark inspired by MT-bench~\cite{mtbench_benchmark}, ChocoLlama-Bench, in which an LLM-as-a-judge (GPT-4o) compares model answers based on $56$ Dutch instructions. Across benchmarks, \textsf{ChocoLlama-2-7B-instruct} and \textsf{ChocoLlama-2-7B-tokentrans-instruct} surpass the original Llama-2 model, effectively demonstrating the use of LoRA at scale for language adaptation. Further, we find that \textsf{ChocoLlama-2-7B-tokentrans-instruct} (the model with a new, Dutch tokenizer) outperforms the model with the original tokenizer \textsf{ChocoLlama-2-7B-instruct} across both benchmarks. This suggests that modifying the tokenizer~\cite{remy2023tik}—especially when paired with careful weight re-initialization— can boost language adaptation.

Lastly, we find that Llama-3~\cite{llama3modelcard}, which was released concurrently with this project, outperforms all the previously considered Dutch LLMs of similar model size. We attribute this to the fact that Llama-3 has been trained on $15$ trillion, multilingual tokens, likely also including all Dutch data that we here consider. We then apply the exact same language adaptation strategy to Llama-3 (using the original tokenizer) to achieve \textsf{Llama-3-ChocoLlama-instruct}. While this model surpasses the Llama-3-instruct in our quantitative benchmark, we find that Llama-3-instruct surpasses \textsf{Llama-3-ChocoLlama-instruct} on ChocoLlama-Bench, which we empirically find to be more reliable. 

The results for Llama-3 suggest that for foundation models with already significant multilingual capabilities, further pretraining on Dutch data yields limited returns. We hypothesize that the favorable results for Llama-3-instruct originate from a better posttraining strategy and  would argue that language adaptation starting from increasingly better, multilingual foundation models should focus more on domain-specific posttraining rather than continued pretraining. However, we note that for adapting Llama-3, we chose to use the original tokenizer rather than a Dutch-specific tokenizer, as we anticipated limited benefits given the substantial vocabulary size of the original model ($128$k tokens for Llama-3 compared to $32$k tokens for Llama-2). We leave for future work whether modifying the tokenizer also improves language adaption (either in terms of language generation quality or efficiency) for models with advanced multilingual capabilities. 

In conclusion, we hope this project contributes to the research of adapting LLMs to lower resource languages as a whole, and to the development of ever improving Dutch LLMs in particular. Our results for successfully teaching Llama-2 Dutch show that (1) the parameter efficient fine-tuning method LoRA\cite{hulora} can be effective at scale and that (2) adaptation of the tokenizer and embedding reinitialization can significantly boost performance in language adaptation. We argue that future work is needed to develop more reliable benchmarks for lower-resource languages and to develop more targeted language adaptation techniques for models with increasing multilingual proficiency. 

We make our models and our new benchmark open-source available on Hugging Face (\url{https://huggingface.co/ChocoLlama}, and provide all code used to generate the results on Github (\url{https://github.com/ChocoLlamaModel/ChocoLlama}).

\section{Pretraining}
\subsection{Data collection}
\label{sec:data}

While we empirically find publicly available Dutch data to be relatively scarce, we were able to collect $104$ GB of Dutch textual data in total, yielding $32$B tokens using the original Llama-2 tokenizer~\cite{touvron2023llama2} (used below to compute the number of tokens unless stated otherwise). To put this in perspective, the original Llama-2 model was pretrained on $2$ trillion tokens, while Geitje-7B~\cite{rijgersberg2023geitje}  was obtained by further pretraining Mistral 7B~\cite{jiang2023mistral} on $10$ billion Dutch tokens. 

The Dutch data we here collect originates from a variety of public sources, which we distinguish as:

\begin{enumerate}
    \item \textbf{OSCAR.} The vast majority of our data stems from the Dutch part of OSCAR~\cite{suarez2019asynchronous}. We here consider its January 2023 version, which is based on the November/December 2022 dump from Common Crawl~\cite{commoncrawl}. This dataset represents $93$ GB of text or $28.6$B tokens. 
    \item \textbf{Open Subtitles.} We collect Dutch data from movie subtitles~\cite{lison2016opensubtitles2016}. Specifically, we consider the text from all unique movies either originally in Dutch or with Dutch subtitles, leading to $214$k data samples. This dataset represents $5$ GB of text or $1.54$B tokens.
    \item \textbf{Project Gutenberg.} We download all Dutch books made available in the public domain on Project Gutenberg~\cite{projectgutenberg}. We use a publicly available scraper~\cite{kpullygutenberg} to download the $970$ full Dutch books available on September 2023. We released this dataset on Hugging Face\footnote{\url{https://huggingface.co/datasets/ChocoLlama/gutenberg-dutch}}. This dataset represents $0.3$ GB of text or $92$M tokens.
    \item \textbf{Wikipedia.} We download the datadump from Wikipedia~\cite{wikidump} from March 2023. We realize that there might be some level of duplication between this data and OSCAR, but given the typically high quality of Wikipedia text we still add the dataset in full. This dataset represents $2.5$ GB of text or $769$M tokens.
    \item \textbf{Job descriptions (TechWolf).} We consider a random sample of $750$k Dutch job descriptions, collected from publicly accessible websites over the past five years and made available to us thanks to TechWolf. This dataset represents $1.5$ GB of text or $462$M tokens. 
    \item \textbf{\textit{Staatsblad} (Bizzy).} We consider a random sample of $80$k of legal filings related to Belgium-based corporations made publicly available by \textit{Het Belgisch Staatsblad}\footnote{\url{https://www.ejustice.just.fgov.be/cgi/welcome.pl}}. We included only documents that could be processed using Optical Character Recognition (OCR) technology and made our best effort to remove any data concerning specific individuals by excluding the relevant sections of the documents. We thank Bizzy for enabling the collection of this dataset. This dataset represents $1.4$ GB of text or $431$M tokens.
    \item \textbf{Legislation (ML6).} We collect a random sample of 15k documents from Flemish legislation using their Open Data API\footnote{\url{https://www.vlaanderen.be/vlaams-parlement/de-vlaamse-codex}}. We thank ML6 for enabling the collection of this dataset. This dataset represents $0.2$ GB of text or $62$M tokens.
\end{enumerate}

While all data is sourced from public websites, we do not release the data collected in cooperation with our partners (TechWolf, Bizzy ML6), in accordance with the licensing terms we agreed upon.

\subsection{Pretraining implementation details}

We here further pretrain three distinct LLMs on our entire Dutch dataset for $1$ epoch.

First, we develop \textsf{ChocoLlama-2-7B-base}, adapting LlaMA-2 7B~\cite{touvron2023llama2} to Dutch using its original tokenizer. The total dataset consists of $32$B tokens. We apply continued pretraining (i.e. updating all the parameters) to both the embedding layers as well as the head of the language model. For all layers in between, we apply the parameter efficient fine-tuning method LoRA~\cite{hulora} to all its available target modules with $r=8$ and $\alpha = 32$. This leads to $544$M trainable parameters, equivalent to $7.75\%$ of all 7B model parameters. 

Second, we adapt the tokenizer to one trained exclusively on Dutch data and develop the model \textsf{ChocoLlama-2-7B-tokentrans-base}. Specifically, we train a BPE tokenizer on the Dutch portion of the OSCAR corpus~\cite{suarez2019asynchronous} (January 2023 dump) with a vocabulary size of $50,000$ tokens. Note that this is larger than the vocabulary size of Llama-2~\cite{touvron2023llama2}, which is $32,000$. We realize that changing a component of the LLM as fundamental as the tokens it considers can lead to complete \emph{catastrophic forgetting} of all the language capabilities and knowledge stored in the original LLM. Hence, we apply the model reinitialization technique introduced by Remy et al.~\cite{remy2023tik}. This technique allows us to map tokens from the Dutch tokenizer to semantically similar tokens from the original, primarily English tokenizer. Initially proposed for masked language models, showing a great performance increase in language adaptation, we now apply this technique to our setup of autoregressive language modeling with Llama-2~\cite{touvron2023llama2}. Using this new tokenizer, the total number of tokens is reduced to $22.6$B, or a reduction of $29.4$\% compared to the original tokenizer. This is expected, as a Dutch-specific tokenizer (and one with a larger vocabulary) would more efficiently encode the same textual data. 
We further consider the same training setup as for \textsf{ChocoLlama-2-7B-base}, which now leads to a slightly higher number of trainable parameters (with the increase in vocabulary size), i.e. an equivalent of $11.47\%$ of all 7.3B parameters, or 839M in total.

Third and lastly, we also develop \textsf{Llama-3-ChocoLlama-base} - applying the same language adaptation strategy to Llama-3 8B~\cite{llama3modelcard} as used for \textsf{ChocoLlama-2-7B-base}. This now leads to 1.07B parameters that are trainable, or an equivalent of $11.77\%$ of all 8B parameters. While this project was initially exclusively focused on Llama-2~\cite{touvron2023llama2}, the new model version was released during the development of \textsf{ChocoLlama} and its great performance motivated us to extend our language adaptation efforts to the latest model. We decided to now consider the original tokenizer, and not the Dutch-specific tokenizer. We opt for this setup as the original Llama-3~\cite{llama3modelcard} model already considers an extensive vocabulary size of $128$k tokens. When comparing the vocabulary of each tokenizer, we find that 30.0\% of all tokens from the Dutch-specific tokenizer are also in the Llama-3 tokenizer, while this was only 7.9\% for Llama-2. We hypothesized that any potential gains from adding Dutch-specific tokens would be now more incremental compared to the risk of catastrophic forgetting, and chose to use the original tokenizer when adapting Llama-3 to Dutch. The total dataset, encoded using the tokenizer of Llama-3, consists of $17$B tokens. This represents a significant reduction compared to the $32$B tokens from the original Llama-2 tokenizer, already suggesting that Llama-3's capabilities might transfer more effectively to Dutch than Llama-2. 

Across all three training runs, we further consider the truncated floating point precision bfloat16 (\texttt{bf16}) for fully sharded and data parallel training on $2$ nodes of each $4$ A100 NVIDIA GPUs (80GB RAM). We use a per-device batch size of $16$ for each of the $8$ GPUs, and consider a distinct number of gradient accumulation steps per model (see Table~\ref{tab:hyperparams}). Across all models we use a context window of $4096$. We use a learning rate of $3e-4$ with a step-based scheduler, multiplying the learning rate by $\gamma=0.85$ every $10\%$ of the entire pretraining dataset. The code used to develop the models described above is largely inspired by the functionalities provided by Meta AI~\cite{llama_recipes}.

\begin{table}[ht]
    \centering
    \caption{Pretraining details to develop \textsf{ChocoLlama-base} models. }
    \begin{tabular}{l|cccc|c|c}
    \toprule
         &  Total & Batch & Total & Gradient & Total & Wall time\\
        Model &  samples & size & steps & acc. steps & Params & (days)\\
        \midrule
        \textsf{ChocoLlama-2-7B} & $7.95$M & $128$ & $62.1$k & $8$ & $544$M & $18.4$\\ 
        \textsf{ChocoLlama-2-7B-tokentrans} & $5.61$M & $128$ & $43.8$k & $6$ & $839$M & $13.1$\\ 
        \textsf{Llama-3-ChocoLlama} & 3.69M & 512 & 57.6k & 8 & 1072M & $14.0$\\ 
    \bottomrule
    \end{tabular}
    \label{tab:hyperparams}
\end{table}

Table~\ref{tab:hyperparams} summarizes the key differences in pretraining across the three models. Although all models are trained on the same Dutch textual dataset, variations in the tokenizer result in a different number of tokenized samples. For instance, the new Dutch tokenizer we here develop leads to 29.4\% less data samples, indicating that the new tokenizer more efficiently compresses the same text to a sequence of tokens. Additionally, differences in model architecture, particularly for Llama-3~\cite{llama3modelcard}, lead to varying numbers of trainable parameters across pretraining runs. We report the final wall time used to complete the pretraining. Due to the specifics of the computing architecture used in this project (provided by the VSC, see Sec \ref{sec:acknowledgements}), each pretraining run was constrained to a 72-hour wall time, requiring the entire pretraining to be executed as a series of consecutive tasks (SLURM jobs).

\subsection{Pretraining results}

\begin{figure*}[!h]
    \centering
    \subfloat[Model perplexity $\mathcal{P}_{\theta}(D_{\text{batch}})$.]{\includegraphics[width=0.45\textwidth]{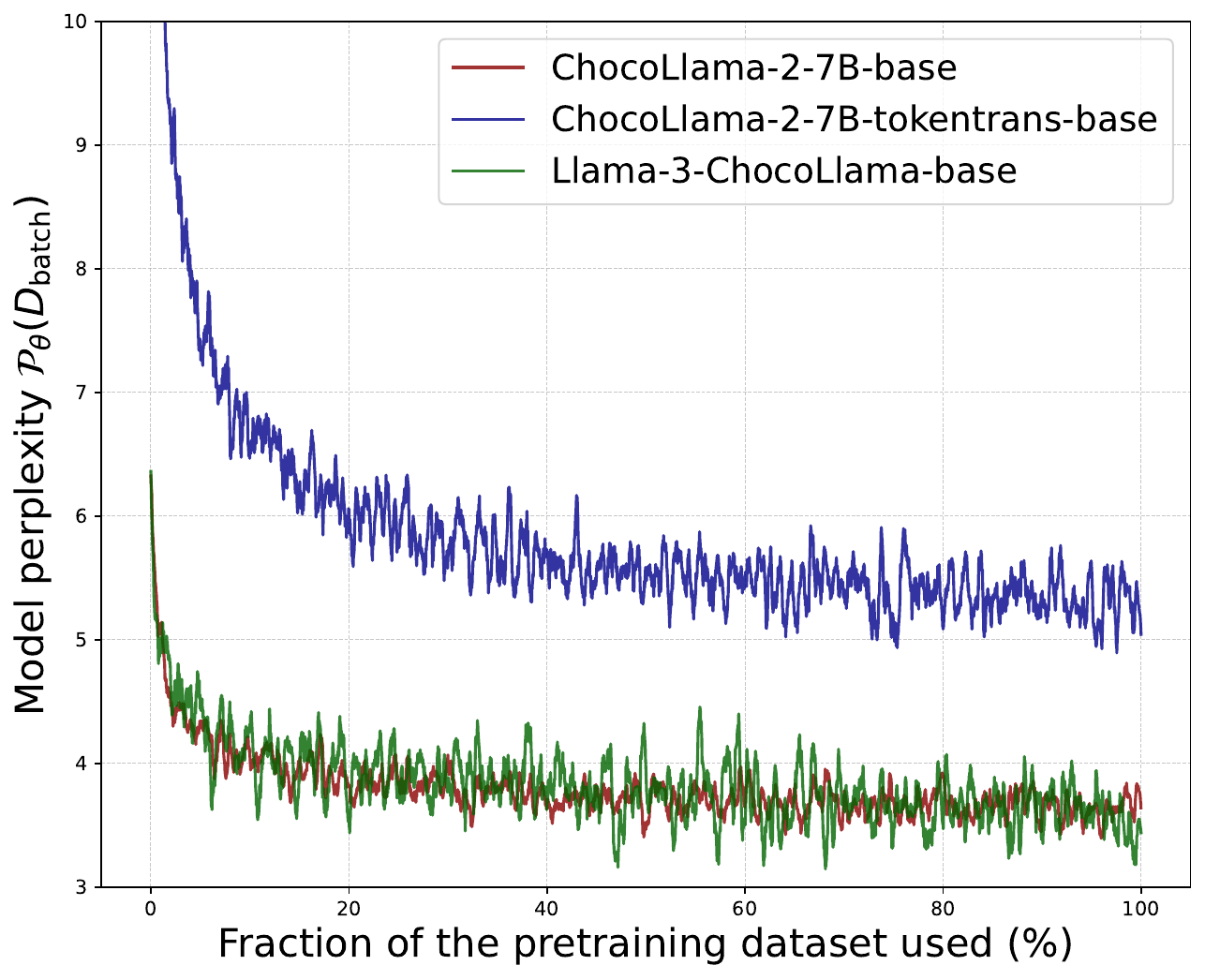}}
    \qquad
    \subfloat[Normalized perplexity $\mathcal{P}_{\theta}^\text{norm}(D_{\text{batch}})$.]{\includegraphics[width=0.45\textwidth]{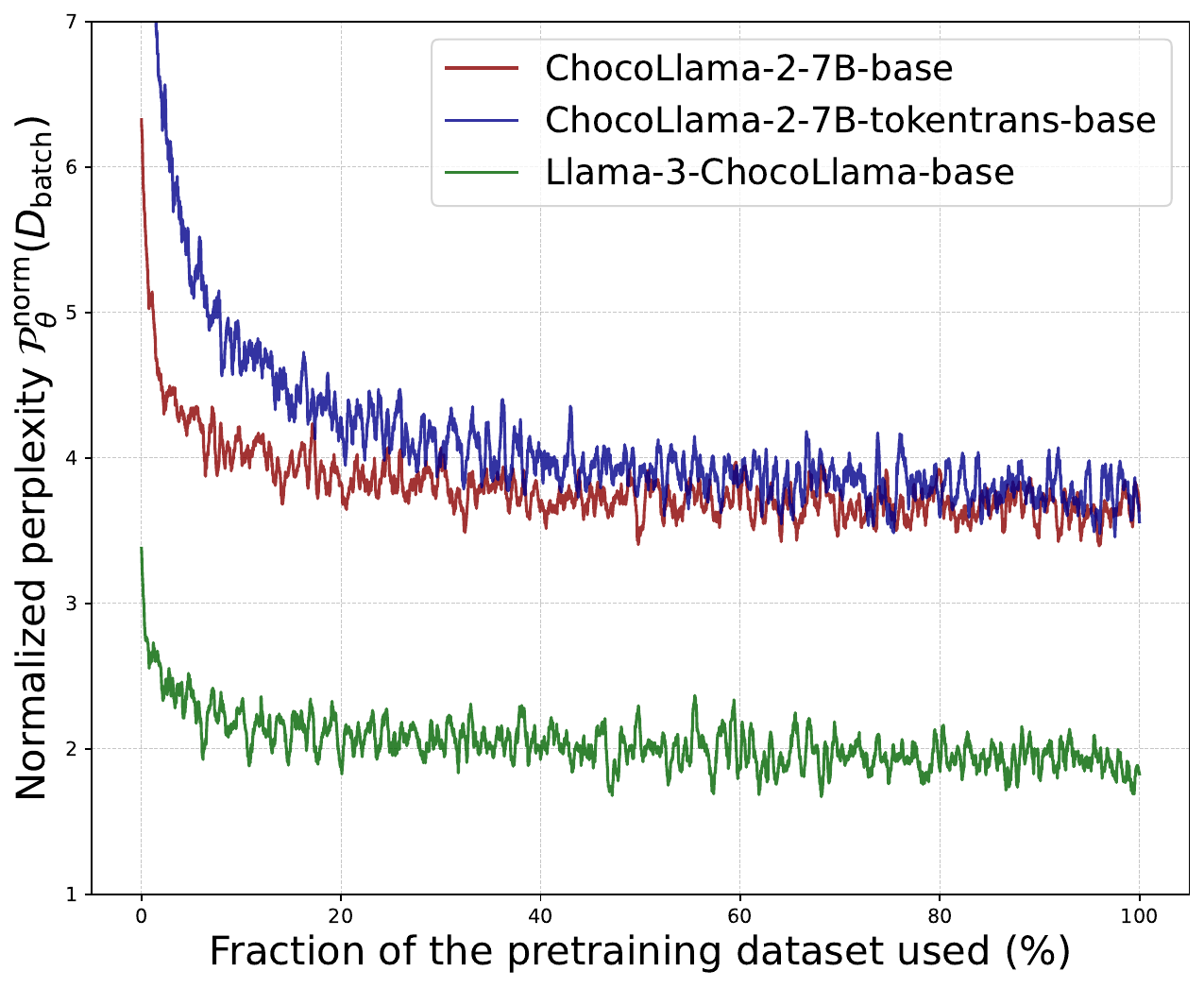}}
    \caption{Model perplexity $\mathcal{P}_{\theta}(D_{\text{batch}})$ (left) and normalized perplexity $\mathcal{P}_{\theta}^\text{norm}(D_{\text{batch}})$ (see Equation \ref{eq_norm_perpl}) (right) across the ChocoLlama model suite during pretraining over 1 full epoch of the collected Dutch data.}
    \label{fig:pretraining}
\end{figure*}

During training, we track the perplexity $\mathcal{P}_{\theta}(D_{\text{batch}})$ computed on the latest (training) $D_{\text{batch}}$ using model checkpoint $\theta$. Perplexity is computed as the exponent of the mean cross-entropy loss used for training. Figure~\ref{fig:pretraining}(a) shows how the training perplexity decreases over the full epoch of all Dutch pretraining data, across all models we here consider. 

While the observed trends in training perplexity provide insights into the models' learning progress, it is important to note that the perplexity values are not directly comparable across models due to differences in their tokenization. As pointed out by Mielke~\cite{mielke2019can}, perplexity is indeed a token-level metric which should be normalized to a common unit before comparing across tokenization schemes. In this work, we normalize perplexity to the number of tokens according to a reference tokenizer, i.e. the original tokenizer of Llama-2~\cite{touvron2023llama2}. Let $\mathcal{T}_{\theta}$ denote the tokenizer of model $\theta$. For a given reference dataset $D_{\text{ref}}$, let $|\mathcal{T}_{\theta}(D_{\text{ref}})|$ represent the number of tokens when the reference dataset is tokenized using $\mathcal{T}_{\theta}$. The normalized perplexity for model $\theta$ is then computed as:

\begin{equation}
\label{eq_norm_perpl}
    \mathcal{P}_{\theta}^\text{norm}(D_{\text{batch}}) = \mathcal{P}_{\theta}(D_{\text{batch}}) \cdot \frac{|\mathcal{T}_{\theta}(D_{\text{ref}})|}{|\mathcal{T}_{\theta_{\text{ref}}}(D_{\text{ref}})|}
\end{equation}

where $\mathcal{P}_{\theta}$ is the perplexity of model $\theta$ as evaluated on $D_{\text{ref}}$ using its native tokenizer $\mathcal{T}_{\theta}$. For $\theta_{\text{ref}}$ we choose Llama-2, while for $D_{\text{ref}}$ we simply use the entire pretraining dataset. 

Figure~\ref{fig:pretraining}(b) illustrates the normalized perplexity throughout training across all models. This normalization now enables a fair comparison of model performance across diverse tokenization schemes. 

First, we note that the original Llama-2 7B~\cite{touvron2023llama2} model starts off with a moderately low normalized perplexity already. This means, without explicitly being trained on Dutch, the model already has a significant performance in the generation of Dutch natural language. When being further pretrained on Dutch data, as done for \textsf{ChocoLlama-2-7B-base}, the perplexity decreases strongly for the first thousands of steps to eventually showing little performance increase closer to the end of the epoch. 

Second, we find that the model with token translation, \textsf{ChocoLlama-2-7B-tokentrans-base}, starts off with a significantly larger normalized perplexity than the original Llama-2 model. This makes sense, as changing the tokenization, despite the reinitialization of the embeddings as proposed by Remy et al.~\cite{remy2023tik}, likely has a significant impact on the model performance. Yet, we observe that after a decent amount of steps, the model recovers and its perplexity steadily decreases. Notably, throughout pretraining the normalized perplexity of \textsf{ChocoLlama-2-7B-tokentrans-base} catches up with the decreasing perplexity of \textsf{ChocoLlama-2-7B-base}, resulting in similar final values for the normalized perplexity at the end of the epoch.

Lastly, we examine the normalized perplexity of the \textsf{Llama-3-ChocoLlama-base}. We find that the normalized perplexity of the original Llama-3 model (i.e. the beginning of the curve) to be significantly lower than the original Llama-2 models, but also lower than the normalized perplexity of both its Dutch-adapted versions at the end of pretraining. This suggests that Llama-3's Dutch capabilities are significantly stronger from the start than for the original Llama-2 model. Throughout pretraining, we find the normalized perplexity to steadily decrease further, in a similar fashion as \textsf{ChocoLlama-2-7B-base}.

\section{Posttraining}

We further optimize the pretrained models for conversational capabilities.

Specifically, we adopt the same strategy as used to align GEITje-7B to GEITje-7B-ultra~\cite{geitjeultra}. First, we apply supervised finetuning (SFT), utilizing the data made available by Vanroy~\cite{vanroy2023language}, including translated versions of databricks-dolly-15k~\cite{DatabricksBlog2023DollyV2} and a cleaned version of the Alpaca dataset~\cite{alpaca} and synthetically generated data using GPT-4~\cite{gpt4techreport}. Next, we apply Direct Preference Optimization (DPO)~\cite{rafailov2024direct} to the SFT version of all the pretrained models we here develop, now utilizing a Dutch version of the data used to train Zephyr-7B-$\beta$~\cite{tunstall2023zephyr}\footnote{\url{https://huggingface.co/datasets/BramVanroy/ultra_feedback_dutch}}. For both the SFT and DPO stage, we apply the same set of parameters to all models as used in GEITje-7B-ultra~\cite{geitjeultra}, including updating all model hyperparameters. Further, we leverage publicly available code~\cite{alignment_handbook2023} and use a set of 4 NVIDIA A100 (80 GB RAM) for both stages. This results in three final models: \textsf{ChocoLlama-2-7B-instruct}, \textsf{ChocoLlama-2-7B-tokentrans-instruct} and \textsf{Llama-3-ChocoLlama-instruct}.

Importantly, as the data used for the finetuning stage has been constructed using GPT-4~\cite{gpt4techreport}, the finetuned models can not be used for commercial purposes. However, we merely apply the finetuning to demonstrate the capabilities of the pretrained models. For any specific use-cases, we still recommend using corresponding Dutch data to finetune the pretrained models, which we do release under a permissive license.

\section{Related work on Dutch language models} 
In recent years, other language models specifically optimized for the Dutch natural language have been developed. We here provide a brief overview of the developments most relevant to our work and identify a series of models to benchmark the Dutch language capabilities of \textsf{ChocoLlama}.

First, prior work has trained masked language models from scratch on Dutch and have showed these to outperform models with multilingual capacity. Delobelle at al.~\cite{delobelle2020robbert} trained a RoBERTa model~\cite{liu2019roberta} on 39GB of Dutch data from the OSCAR corpus~\cite{suarez2019asynchronous} and achieved state-of-the-art downstream performance compared to the multilingual mBERT~\cite{devlin2018bert}. Interestingly for this work, authors consider both pretraining with the original BPE tokenizer and its vocabulary as well as training a new tokenizer on Dutch data, concluding that also adapting the tokenizer to the target language leads to improved performance. Concurrently, two other Dutch masked language models were developed, BERT-NL~\cite{brandsen2019bert} and BERTje~\cite{de2019bertje}, both also finding increased performance compared to multilingual models.  

Further research has investigated methods for "translating" tokenizers from the original language to a target language by re-initializing the original model's embeddings based on semantic similarities between tokens. Remy et al.\cite{remy2023tik} demonstrated that this approach enables efficient language adaptation for BERT-based~\cite{devlin2018bert} masked language models, a technique we leveraged to develop \textsf{ChocoLlama-2-7B-tokentrans-base}. Concurrently with this work, a similar strategy was employed to create tweety-7b-dutch~\cite{remy2024transtokenization}, a trans-tokenized adaptation of Mistral 7B~\cite{jiang2023mistral} to Dutch.

Most relevant to our work, autoregressive LLMs have also been finetuned to Dutch. GEITje-7B~\cite{rijgersberg2023geitje} is a full-parameter finetuned model starting from Mistral 7B~\cite{jiang2023mistral} further trained on 10 billion tokens of Dutch text. Further versions of GEITje-7B have been developed, optimized for conversational behaviour leveraging synthetic (translated) instructions and conversations made available by Vanroy~\cite{vanroy2023language}. The best performing model of this suite is reportedly GEITje-7B-ultra~\cite{geitjeultra}, which applies direct preference optimization~\cite{rafailov2024direct} on a supervised finetuned version of the base model. Further, models have also been developed without continued pretraining on Dutch data and just leveraging Dutch instruction datasets. Recent examples include  Reynaerde-7B-Chat~\cite{reynaerdechat}, a QLoRA~\cite{dettmers2024qlora} finetuned version of the instruction-tuned version of Mistral 7B v0.3~\cite{jiang2023mistral}, and ReBatch/Llama-3-8B-dutch~\cite{llama3bdutch}, a QLORA~\cite{dettmers2024qlora} and ORPO~\cite{hong2024orpo} fine-tuned version of Llama-3-8B~\cite{llama3modelcard}. 

Finally, a leaderboard has been established to track and compare the Dutch language capabilities of newly developed LLMs\footnote{\url{https://huggingface.co/spaces/BramVanroy/open_dutch_llm_leaderboard}}. 

\section{Evaluation}
\label{sec:evaluation}

\subsection{Quantitative evaluation using readily available Dutch benchmarks} 

We first evaluate our models using Dutch translations (using GPT 3.5 turbo) of industry-standard benchmarks, provided by the Open Multilingual LLM Evaluation Leaderboard~\cite{lai2023openllmbenchmark}: ARC (25-shot)~\cite{arc_benchmark}, HellaSwag (10-shot)~\cite{hellaswag_benchmark}, MMLU (5-shot)~\cite{mmlu_benchmark} and TruthfulQA (0-shot)~\cite{truthfulqa_benchmark}. The results for all of our \textsf{ChocoLlama} models, both after pretraining and posttraining, compared with prior work, can be found in Table~\ref{tab:quantitative-results}. We provide the results for each benchmark dataset (accuracy in multiple choice questions) as well as an average performance across benchmarks, which is used to rank all models from high to low performance. 

First, we compare the performance of the pretrained models which have not been subjected to any instruction tuning. We find both \textsf{ChocoLlama-2-7B-base} and \textsf{ChocoLlama-2-7B-tokentrans-base} to outperform its original base model Llama-2-7b-hf~\cite{touvron2023llama2}. This confirms that applying continued pretraining exclusively on the Dutch data we have collected leads to improved Dutch language capabilities. In contrast, we find a slight decrease in performance for \textsf{Llama-3-ChocoLlama-base} compared to Meta-Llama-3-8B~\cite{llama3modelcard}. 

Further, we find the model with a Dutch-specific tokenizer, \textsf{ChocoLlama-2-7B-tokentrans-base}, to slightly outperform the model with the original tokenizer, \textsf{ChocoLlama-2-7B-base}. This suggests that adapting the tokenizer specifically to Dutch -with clever weight reinitialization as proposed by Remy et al.~\cite{remy2023tik}- provides a promising approach for language adaptation. While the initial model perplexity when changing the tokenizer increased significantly compared to the original model (see Figure~\ref{fig:pretraining}(b)), the continued pretraining on Dutch data seems to have largely mitigated any catastrophic forgetting, leading to a final performance even slightly better than achieved with the original tokenizer. Notably, a tokenizer specific to Dutch is also more efficient in compressing Dutch language, and would considerably reduce inference cost compared to the models with the original tokenizer. 

As expected, we also find that the instruction-tuned models consistently outperform their base models, reaffirming the importance of instruction-tuning, especially for performance measured on these evaluation benchmarks. Particularly, we find \textsf{ChocoLlama-2-7B-tokentrans-instruct} to outperform all Llama-2 based models. This confirms that the language adaptation strategy we here apply has improved Llama-2's capabilities in Dutch. 

However, we still find concurrently developed models, such as Geitje-7b-ultra~\cite{geitjeultra} and Reynaerde-7B-Chat~\cite{reynaerdechat}, to outperform \textsf{ChocoLlama-2-7B-tokentrans-instruct} on average for these benchmarks. We attribute this mostly to the better baseline performance of the foundation model used for language adaptation, as Mistral-7b-v0.1~\cite{jiang2023mistral} significantly outperforms Llama-2-7b-hf~\cite{touvron2023llama2}. Similarly, we observe that for these benchmarks, Meta-Llama-3-8B-Instruct~\cite{llama3modelcard} outperforms all previously developed models. This suggests that Llama-3, with its $15$ trillion tokens used for pretraining, has gained significant multilingual capabilities, including for Dutch. 

Finally, we find that our language adaptation strategy applied to Meta-Llama-3-8B~\cite{llama3modelcard}, resulting in \textsf{Llama-3-ChocoLlama-8B-instruct}, outperforms the previous state-of-the-art in its weight class by moderate margins. 

\begin{table}[t]
\centering
\caption{Quantitative evaluation of \textsf{ChocoLlama} models (highlighted in gray) and prior work on readily available Dutch benchmarks (as implemented by the Open Dutch LLM Leaderboard). We compute the accuracy achieved for multiple choices questions on Dutch translations (using GPT 3.5 turbo) of ARC (25-shot)~\cite{arc_benchmark}, HellaSwag (10-shot)~\cite{hellaswag_benchmark}, MMLU (5-shot)~\cite{mmlu_benchmark} and TruthfulQA (0-shot)~\cite{truthfulqa_benchmark}. Models are ranked based on their average performance across all benchmarks.}
\begin{tabular}{l|cccc|c}
\toprule
& \multicolumn{4}{c}{Evaluation benchmark} & \\
Model & ARC & HellaSwag & MMLU & TruthfulQA & Avg.\\ 
\midrule
\rowcolor{mygray} 
\textbf{Llama-3-ChocoLlama-8B-instruct} & \bfseries 0.48 & \bfseries 0.66 & \bfseries 0.49 & \bfseries 0.49 & \bfseries 0.53 \\
ReBatch/Llama-3-8B-dutch \cite{llama3bdutch} & 0.44 & 0.64 & 0.46 & 0.48 & 0.51\\
Meta-Llama-3-8B-Instruct \cite{llama3modelcard} & 0.47 & 0.59 & 0.47 & 0.52 & 0.51 \\
Meta-Llama-3-8B \cite{llama3modelcard} & 0.44 & 0.64 & 0.47 & 0.45 & 0.5 \\
Reynaerde-7B-Chat \cite{reynaerdechat} & 0.44 & 0.62 & 0.39 & 0.52 & 0.49 \\
\rowcolor{mygray} 
Llama-3-ChocoLlama-base & 0.45 & 0.64 & 0.44 & 0.44 & 0.49\\
Zephyr-7b-beta \cite{tunstall2023zephyr} & 0.43 & 0.58 & 0.43 & 0.53 & 0.49\\
Geitje-7b-ultra \cite{geitjeultra} & 0.40 & 0.66 & 0.36 & 0.49 & 0.48\\
\rowcolor{mygray} 
ChocoLlama-2-7B-tokentrans-instruct & 0.45 & 0.62 & 0.34 & 0.42 & 0.46\\
Mistral-7b-v0.1 \cite{jiang2023mistral} & 0.43 & 0.58 & 0.37 & 0.45 & 0.46\\
\rowcolor{mygray} 
ChocoLlama-2-7B-tokentrans-base & 0.42 & 0.61 & 0.32 & 0.43 & 0.45\\
\rowcolor{mygray} 
ChocoLlama-2-7B-instruct & 0.36 & 0.57 & 0.33 & 0.45 & 0.43\\
\rowcolor{mygray} 
ChocoLlama-2-7B-base & 0.35 & 0.56 & 0.31 & 0.43 & 0.41\\
Llama-2-7b-chat-hf \cite{touvron2023llama2} & 0.36 & 0.49 & 0.33 & 0.44 & 0.41\\
Llama-2-7b-hf \cite{touvron2023llama2} & 0.36 & 0.51 & 0.32 & 0.41 & 0.40\\
\bottomrule
\end{tabular}
\label{tab:quantitative-results}
\end{table}

\subsection{Qualitative evaluation using ChocoLlama-Bench}

\subsubsection{ChocoLlama-Bench as a new benchmark for Dutch language models} 

While the multiple-choice format of the benchmarks mentioned above provides a reasonable way to evaluate a model's language comprehension, it may not be ideal for assessing a model's ability to \emph{generate} Dutch text. Since our primary objective is to achieve a model capable of generating Dutch language, we here also propose a qualitative benchmark in which we compare side-by-side multi-turn conversations generated by two models. To this end, we have compiled a diverse set of $30$ Dutch prompts and follow-up prompts (in the style of MT-bench~\cite{mtbench_benchmark}) for a total of $56$ turns. The prompts span across three categories: reasoning, comprehension and writing. We implemented ChocoLlama-Bench using the FastChat repository~\cite{fastchat_repo}, and provide the necessary resources for reproduction in the code we release. For an overview of the prompts, please refer to the ChocoLlama-Bench dataset on HuggingFace\footnote{https://huggingface.co/datasets/ChocoLlama/ChocoLlama-Bench}. 

We query a range of models from Table~\ref{tab:quantitative-results} to answer the prompts from ChocoLlama-Bench. We then use GPT-4o~\footnote{https://openai.com/index/hello-gpt-4o/} to compare the performance of two models side-by-side and judge which model's answer was preferred. In the instructions to the evaluation model, we specify that it should assess responses based on helpfulness, relevance, accuracy, depth, creativity, and level of detail. We also explicitly state that for a model's answer to be preferred it needs to be in Dutch (and not in English). We provide the number of times each model wins per task (or the number of ties) per category. 

\subsubsection{Teaching Llama-2 Dutch} 

First, we analyze the two main, instruction-tuned models we developed based on Llama-2-7B~\cite{touvron2023llama2}: \textsf{ChocoLlama-2-7B-instruct} and \textsf{ChocoLlama-2-7B-tokentrans-instruct}. Using our evaluation model, we assess whether answers from these ChocoLlama variants are preferred over those from the instruction-finetuned version of the original Llama-2-7B model. The results in Table~\ref{tab:qualitative-results-chocollama-2} show that \textsf{ChocoLlama-2-7B-instruct} and \textsf{ChocoLlama-2-7B-tokentrans-instruct} outperform Llama-2 in 78\% and 66\% of the cases, respectively. A closer inspection of the responses reveals that while Llama-2 primarily provides answers in English, the ChocoLlama models consistently respond in Dutch. For specific examples, please refer to Appendix \ref{app:samples}. These findings indicate that we have successfully adapted an originally English model to effectively understand and generate responses in Dutch.

We then also directly compare our two models, \textsf{ChocoLlama-2-7B-instruct} with the original Llama-2 tokenizer and \textsf{ChocoLlama-2-7B-tokentrans-instruct} with its new, Dutch-specific tokenizer, to measure the impact of tokenizer translation on performance. The model with the new tokenizer—\textsf{ChocoLlama-2-7B-tokentrans-instruct}—demonstrates a slight edge, winning in 43\% of the tasks. In particular, for the questions related probing the model's writing capabilities, we find that the \textsf{ChocoLlama-2-7B-tokentrans-instruct} has the largest advantage. This suggests that for primarily English models, updating the tokenizer prior to fine-tuning may enhance adaptation to specific domains or languages.

\begin{table}[h]
\centering
\caption{Qualitative side-by-side evaluation of \textsf{ChocoLlama} (highlighted in gray) instruction-tuned models based on Llama-2-7B on ChocoLlama-Bench. We query each model on a total of $56$ custom Dutch prompts across categories (Reasoning, Comprehension, Writing) and report which answer the judgement model (GPT-4o) prefers. The model which wins the most turns is highlighted in bold.}
\begin{tabular}{l|ccc|c}
\toprule
Winning model & Reasoning & Comprehension & Writing & Total \\ 
\midrule
\midrule
\rowcolor{mygray} 
\textbf{ChocoLlama-2-7B-instruct} & 16 & 10 & 18 & \textbf{44 (78\%)} \\
Tie & 1 & 1 & 4 & 6 (11\%) \\
Llama-2-7b-chat & 1 & 3 & 2 & 6 (11\%) \\
\midrule
\midrule
\rowcolor{mygray} 
\textbf{ChocoLlama-2-7B-tokentrans-instruct} & 13 & 8 & 16 & \textbf{37 (66\%)} \\
Tie & 5 & 5 & 7 & 17 (30\%) \\
Llama-2-7b-chat & 0 & 1 & 1 & 2 (4\%) \\
\midrule
\midrule
\rowcolor{mygray} 
\textbf{ChocoLlama-2-7B-tokentrans-instruct} & 8 & 4 & 12 & \textbf{24 (43\%)} \\
Tie & 5 & 6 & 6 & 17 (30\%) \\
\rowcolor{mygray} 
ChocoLlama-2-7B-instruct & 5 & 4 & 6 & 15 (27\%) \\
\midrule
\bottomrule
\end{tabular}
\label{tab:qualitative-results-chocollama-2}
\end{table}

\subsubsection{Comparison to prior work} 

We now compare how our best performing model based on Llama-2 compares to models adapted to Dutch developed in prior and concurrent work. We compare \textsf{ChocoLlama-2-7B-tokentrans-instruct} to two instruction-tuned models GEITje-7B-ultra~\cite{geitjeultra} and Reynaerde-7B-Chat~\cite{reynaerdechat} - both based on finetuning of Mistral-7B~\cite{jiang2023mistral}. Table~\ref{tab:qualitative-results-competition} summarizes the results.  

Consistent with the results from the benchmarks in Table~\ref{tab:quantitative-results}, we find that both GEITje-7B-ultra~\cite{geitjeultra} and Reynaerde-7B-Chat~\cite{reynaerdechat} outperform \textsf{ChocoLlama-2-7B-tokentrans-instruct} winning for 73\% and 37\% of the turns. We hypothesize this to be due to superior multilingual --and Dutch-- language capabilities from Mistral-7B as the base model compared to Llama-2 (see Table~\ref{tab:quantitative-results}). 

\begin{table}[h]
\centering
\caption{Qualitative side-by-side evaluation of \textsf{ChocoLlama} (highlighted in gray) instruction-tuned models on ChocoLlama-Bench, compared to Dutch instruction-tuned models developed by prior and concurrent work. We query each model on a total of $56$ custom Dutch prompts across categories (Reasoning, Comprehension, Writing) and report which answer the judgement model (GPT-4o) prefers. The model which wins the most turns is highlighted in bold.}
\begin{tabular}{l|ccc|c}
\toprule
Winning model & Reasoning & Comprehension & Writing & Total \\ 
\midrule
\midrule
\rowcolor{mygray} 
ChocoLlama-2-7B-tokentrans-instruct & 1 & 0 & 4 & 5 (9\%) \\
Tie & 4 & 3 & 3 & 10 (18\%) \\
\textbf{GEITje-7B-ultra} & 13 & 11 & 17 & \textbf{41 (73\%)} \\
\midrule
\midrule
\rowcolor{mygray} 
ChocoLlama-2-7B-tokentrans-instruct & 4 & 3 & 8 & 15 (27\%) \\
Tie & 4 & 6 & 10 & 20 (36\%) \\
\textbf{Reynaerde-7B-Chat} & 10 & 5 & 6 & \textbf{41 (37\%)} \\
\midrule
\bottomrule
\end{tabular}
\label{tab:qualitative-results-competition}
\end{table}

\subsubsection{Comparison to Llama-3 based models} 

During the development of \textsf{ChocoLlama-2}, Meta-Llama-3-8B-Instruct~\cite{llama3modelcard} was released as well. Table~\ref{tab:qualitative-results-chocollama3} compares the Dutch language capabilities of our best performing Llama-2-based model with the most recent Meta-Llama-3-8B-Instruct. We find that Meta-Llama-3-8B-Instruct wins 57\% of the turns, suggesting that Llama-3 has significant multilingual and Dutch capabilities, better than a Dutch-optimized version of Llama-2. 

We then apply the same language adaptation strategy (maintaining the tokenizer) to develop \textsf{Llama-3-ChocoLlama-8B-instruct}, starting from the pretrained Llama-3 model. We find, in contrast with the benchmarks from Table~\ref{tab:quantitative-results}, that also here Meta-Llama-3-8B-Instruct outperforms our Dutch-specific model on ChocoLlama-bench. 

\begin{table}[h]
\centering
\caption{Qualitative side-by-side evaluation of \textsf{ChocoLlama} (highlighted in gray) instruction-tuned models based on Llama-3-8B on ChocoLlama-Bench. We query each model on a total of $56$ custom Dutch prompts across categories (Reasoning, Comprehension, Writing) and report which answer the judgement model (GPT-4o) prefers. The model which wins the most turns is highlighted in bold.}
\begin{tabular}{l|ccc|c}
\toprule
Winning model & Reasoning & Comprehension & Writing & Total \\ 
\midrule
\midrule
\rowcolor{mygray} 
ChocoLlama-2-7B-tokentrans-instruct & 0 & 1 & 6 & 7 (13\%) \\
Tie & 4 & 3 & 10 & 17 (30\%) \\
\textbf{Meta-Llama-3-8B-Instruct} & 14 & 10 & 8 & \textbf{32 (57\%)} \\
\midrule
\midrule
\rowcolor{mygray} 
Llama-3-ChocoLlama-8B-instruct & 2 & 3 & 4 & 9 (16\%) \\
Tie & 4 & 4 & 12 & 20 (36\%) \\
\textbf{Meta-Llama-3-8B-Instruct} & 12 & 7 & 8 & \textbf{27 (48\%)} \\
\midrule
\midrule
\rowcolor{mygray} 
Llama-3-ChocoLlama-8B-instruct & 1 & 1 & 3 & 5 (9\%) \\
Tie & 7 & 2 & 8 & 17 (30\%) \\
\textbf{ReBatch/Llama-3-8B-dutch} & 10 & 11 & 13 & \textbf{34 (61\%)} \\
\midrule
\bottomrule
\end{tabular}
\label{tab:qualitative-results-chocollama3}
\end{table}

We believe this lack of improvement to be due two main factors. First, the Llama-3 has been pretrained on 15 trillion tokens, likely already containing most of the Dutch tokens we here consider. We would thus expect the impact of further pretraining on Dutch data to be significantly smaller than for the Llama-2 models. Secondly, and we hypothesize this to be the most important factor, the instruction tuning applied to Meta-Llama-3-8B-Instruct is likely to be significantly better than the one we here adopt from prior work. 

We further confirm this by comparing \textsf{Llama-3-ChocoLlama-8B-instruct} to ReBatch/Llama-3-8B-dutch~\cite{llama3bdutch}, finding the latter to win 61\% of the times. As ReBatch/Llama-3-8B-dutch has been developed starting from the instruction-finetuned Meta-Llama-3-8B-Instruct, the model already benefits from the posttraining applied to Meta-Llama-3-8B-Instruct. The model then only becomes better when further posttrained on Dutch data. These results suggest that for ever improving, multilingual, instruction-tuned models, language adaptation might be more effective when focused on domain-specific posttraining rather than continued pretraining.

\section{Discussion}

Our work on adapting Llama-based LLMs to Dutch provides several valuable insights about the challenges and opportunities in language adaptation, particularly for lower-resource languages.

\paragraph{Successful language adaptation for Llama-2.}  Our approach to adapt the primarily English Llama-2 to Dutch proved successful. To our knowledge, parameter-efficient finetuning methods such as LoRA\cite{hulora}, are not typically applied to the scale used in this project. Specifically, we apply LoRA during continued pretraining on 32B Dutch tokens, just updating around 10\% of the overall model parameters to teach a primarily English model Dutch. Our results demonstrate that efficient parameter updates can effectively adapt a model to a new language. 

Our experiments also indicate that modifying the tokenizer can significantly enhance performance across benchmarks. Replacing the tokenizer with one fully trained on Dutch, combined with the embedding initialization strategy suggested by Remy et al.\cite{remy2023tik}, and continued pretraining using LoRA, resulted in \textsf{ChocoLlama-2-7B-tokentrans-instruct} surpassing \textsf{ChocoLlama-2-7B-instruct} on ChocoLlama-Bench. We find this especially meaningful given the inherent risk of catastrophic forgetting when changing the tokenizer, as exemplified by the large initial perplexity during training (see Figure \ref{fig:pretraining}(b)). These findings suggest that, for adapting models to domains or languages that differ substantially from the original training data, modifying the tokenizer—especially when paired with careful weight reinitialization—can be a promising approach. Moreover, we find that changing the tokenizer more efficiently compresses the same amount of Dutch text in 29.4\% less tokens, making it more efficient for language adaptation and less costly at inference time.  

Finally, we observed that the posttraining phase on limited Dutch data effectively enabled conversational behavior in the model.

\paragraph{Developing state-of-the-art LLMs.} Our work also revealed the increasing challenges in developing state-of-the-art models for specific languages or domains. While our adaptation technique worked well with Llama-2, we found ourselves unable to surpass concurrent work based on Mistral-7b, which possessed stronger multilingual capabilities from the start. More strikingly, during our development process, the release of Llama-3 demonstrated performance that exceeded our Dutch-specific model, highlighting the rapid pace of advancement in multilingual capabilities of foundation models.

We argue that these findings lead to an important observation about the relative value of continued pretraining versus posttraining. When we applied our adaptation approach to Llama-3, which already possessed strong Dutch language capabilities, we found that our continued pretraining on Dutch data followed by posttraining did not outperform Meta's instruction-tuned Llama-3 model. This suggests that when the base model already has significant capabilities in the target domain or language, the impact of continued pretraining may be less significant than high-quality instruction tuning.

\paragraph{Benchmarking for lower-resource languages.} Lastly, our experience highlights the complexities of benchmarking language models in lower-resource languages. Although we initially used existing benchmarks for Dutch, we empirically observed that these benchmarks may not consistently or accurately assess a model's ability to generate fluent Dutch text. To address this, we introduced ChocoLlama-Bench as a new benchmark focused on evaluating Dutch language generation. We empirically find this benchmark to offer a more comprehensive and reliable evaluation of Dutch language capabilities, and we hope its open-source availability will facilitate robust benchmarking of future models. 

However, we note that our benchmark assumes the availability of evaluation models with strong proficiency in the target language. We here use GPT-4o for assessment, which may introduce its own biases and limitations. This underscores, more broadly, the challenge of developing reliable evaluation metrics for language models in lower-resource languages.

\section{Conclusion}

The successful language adaptation for Llama-2 teaches us that parameter-efficient finetuning methods at scale enable language adaption. Further, when the target domain is sufficiently distinct from the original training data, switching the tokenizers with embedding reinitialization could offer a promising approach. 

Looking ahead, as foundation models become increasingly multilingual, the focus of language adaptation may need to evolve as well. Future efforts could benefit from targeted adaptations focusing on specific linguistic nuances or cultural elements, likely during posttraining, rather than broadly focused language adaptation during continued pretraining. 

Beyond language adaptation, our findings might also apply to the adaptation of LLMs to specialized domains (e.g. biomedical or legal text). While large models trained on vast amounts of public data are likely to outperform domain-specific models for general use cases, some level of adaptation (during pretraining or by incorporating expert feedback in posttraining) will likely enhance the model's performance in more specialized use cases. We hypothesize that such improvements might even be achievable with smaller models. Although large models are typically the first choice due to their capacity to store vast knowledge, smaller models, when adapted effectively, might also perform well in niche tasks while being more cost-effective.

Lastly, we emphasize that the development of more robust and comprehensive evaluation frameworks for non-English languages remains essential for advancing model accuracy and relevance in diverse contexts.

\section*{Acknowledgements}
\label{sec:acknowledgements}

We thank the `Vlaams Supercomputer Centrum'~\footnote{\url{https://www.vscentrum.be/}} for extending us a Tier-1 Compute grant, providing the computational resources required to complete this project. We also thank the researchers who advised us during the course of the project: Els Lefever, Veronique Hoste, Cédric Goemaere, Karel D'Oosterlinck, Loic De Langhe, Pranaydeep Singh and Bram Vanroy. Finally, we also thank our industry partners TechWolf, Bizzy and ML6 for assisting us to collect specific Dutch data (see Sec~\ref{sec:data}). 

\bibliographystyle{plainnat}
\bibliography{references} 

\newpage
\appendix

\section{Sample model answers}
\label{app:samples}

To compare the performance in Dutch language generation across all models considered in this work, we also provide sample answers generated for a hand-picked prompt. We use as a prompt: 

\begin{enumerate}
    \item \textbf{System prompt: } \textit{"Je ben een expert in Vlaamse cultuur en je antwoordt behulpzaam, in het Nederlands, op de vraag van de gebruiker."}
    \item \textbf{User prompt: } \textit{"Jacques brel, Willem Elsschot en Jan Jambon zitten op café. Waarover zouden ze babbelen?"}
\end{enumerate}

We further generate a maximum of $256$ tokens, and use a temperature of $0.6$ and top-p of $0.9$ for generation. Tables \ref{tab:examples-chocollama-2} and \ref{tab:examples-other} summarize the responses across models. 

While these results are primarily anecdotal and do not allow for any generalizable conclusions, we still deem it insightful to analyze the generated text. First, we find that Llama-2-7b-chat does generate text in Dutch following the provided instructions, while being primarily trained on English data. However, we find that the generated text includes English phrases (e.g. \textit{"Hey there"}) and elaborates on topics (e.g. \textit{"Kurt Weill"}), unrelated to the given prompt. In contrast, both ChocoLlama models based off Llama-2 are able to answer the prompt in consistent and grammatically correct Dutch, while referencing the Belgian public figures in the prompt. We, however, note that both models do not succeed in appropriately associating all three figures in their correct context (e.g. Jacques Brel and Jan Jambon are arguably not most well known as authors). Further, we find the models in Table \ref{tab:examples-other} to more adequately answer the prompt, in line with the benchmark results in Section \ref{sec:evaluation}. Notably, apart from Meta-Llama-3-8B-Instruct, all models associate the Belgian public figures with their correct context. 

\begin{table}[ht]
    \centering
    \caption{Sample responses to a custom Dutch-specific prompt across models considered in this work.}
    \begin{tabular}{p{2cm}|p{12.5cm}}
    \toprule
        Model & Answer \\
        \midrule
        \midrule
        Llama-2-7b-chat & \textit{Ah, een leuk scenario! Jacques Brel, Willem Elsschot en Jan Jambon zitten op een café en babbelen over hun favoriete thema's. Hier zijn enkele mogelijke conversatiepunten: Jacques Brel: "Hey there, Willem! Ik heb het laatste album van Kurt Weill gehoord en het is geweldig! Hij heeft zulk een unieke stijl en zijn liedjes zijn altijd zo emotioneel. Wat ben je ervan van?" Willem Elsschot: "Ah, Kurt Weill? Ja, hij is een geweldig componist. Maar ik moet ook nog even over mijn laatste reis vertellen. Ik was naar Parijs geweest en het was een echte feestviering. De Franse cultuur is zo rijk en interessant." Jan Jambon: "Ja, Parijs is een echte stad met een rijke cultuur. Maar ik wil liever over de Vlaamse literatuur praten...} \\
        \midrule
        ChocoLlama-2-7B-instruct & \textit{Ze zouden waarschijnlijk praten over literatuur, kunst en de hedendaagse Vlaamse samenleving. Het café is een ideale plaats voor gesprekken over de Vlaamse cultuur, en de drie schrijvers zouden samen kunnen reflecteren op hun eigen werk, de invloed van de tijd op de Vlaamse literatuur, en de hedendaagse literaire trends.} \\
        \midrule
        ChocoLlama-2-7B-tokentrans-instruct & \textit{Als de drie schrijvers - Jacques Brel, Willem Elsschot en Jan Jambon - op café zouden zitten, zou het waarschijnlijk een geanimeerde en interessante conversatie zijn over een breed scala aan onderwerpen. Hun persoonlijke ervaringen, literaire invloeden en de wereld om hen heen zouden ongetwijfeld aan bod komen. Ze zouden waarschijnlijk ook filosoferen over de betekenis van kunst, literatuur en het leven, en de impact die hun werken hebben op de wereld.} \\ 
        \midrule
         \bottomrule
    \end{tabular}
    \label{tab:examples-chocollama-2}
\end{table}

\begin{table}[ht]
    \centering
    \caption{Sample responses to a custom Dutch-specific prompt across models considered in this work.}
    \begin{tabular}{p{2cm}|p{12.5cm}}
    \toprule
        Model & Answer \\
        \midrule
        \midrule
        GEITje-7B-ultra & \textit{In een fictief café waar Jacques brel, Willem Elsschot en Jan Jambon samenkomen, zouden ze waarschijnlijk een breed scala aan onderwerpen bespreken, gezien hun diverse achtergronden en bijdragen aan de Vlaamse cultuur. Hier zijn enkele mogelijke gespreksonderwerpen: 1. **Muziek en Poëzie**: Jacques brel, bekend om zijn Franse chansons, zou wellicht verhalen delen over zijn inspiraties en de evolutie van zijn muziek. Willem Elsschot, een gerespecteerde schrijver, zou wellicht zijn gedachten delen over literaire technieken of de impact van zijn werk op de Vlaamse literatuur. 2. **Politiek en Cultuur**: Jan Jambon, als politicus, zou waarschijnlijk de politieke situatie in Vlaanderen en de rol van cultuur daarin bespreken...}\\
        \midrule
        Reynaerde-7B-Chat & \textit{Jacques Brel, Willem Elsschot en Jan Jambon, drie beroemde figuren uit het Belgische culturele landschap, zouden waarschijnlijk verschillende onderwerpen bespreken. Jacques Brel, bekend om zijn emotionele en introspectieve liedjes, zou het waarschijnlijk hebben over zijn artistieke processen, de invloed van zijn persoonlijke ervaringen op zijn werk, en misschien zou hij ook kritiek uiten op de maatschappelijke kwesties van die tijd. Willem Elsschot, een invloedrijke schrijver, zou het hebben over literaire thema's, de rol van de auteur in de samenleving, en zijn eigen schrijfproces. Jan Jambon, een prominente politicus en schrijver, zou het hebben over politieke kwesties, de relatie tussen kunst en politiek, en mogelijk hun gedeelde interesse in het Frans-Vlaamse culturele erfgoed...} \\  
        \midrule
        Meta-Llama-3-8B-Instruct & \textit{Wat een leuke combinatie! Jacques Brel, Willem Elsschot en Jan Jambon, drie Vlaamse cultuuriconen van verschillende disciplines, zitten samen op café. Hier een mogelijk gesprek: Jacques Brel: "Hé, Jan, hoe gaat het met je nieuwe boek? Heb je al wat kritiek gekregen?" Jan Jambon: "Ah, Jacques, het is een beetje een chaos. Sommige mensen vinden het te abstract, anderen vinden het te concreet. Maar ik ben er trots op dat ik mijn eigen weg ga." Willem Elsschot: "Ik ken dat gevoel. Mijn roman 'Vlaamse Filosofie' was ook niet meteen een bestseller, maar nu wordt het toch erkend als een klassieker."Jacques Brel: "Dat is waar, Willem. Je had een eigen stijl, een eigen manier van schrijven. En ik denk dat Jan hier ook zijn eigen weg gaat. We moeten ons niet bekommeren om wat anderen zeggen, maar ons eigen werk maken."}  \\  
        \midrule
        Llama-3-ChocoLlama-8B-instruct & \textit{Jacques Brel, met zijn muzikale flair en levenslust, zou waarschijnlijk vertellen over zijn avonturen op het podium en de mensen die hij heeft ontmoet. Willem Elsschot, met zijn scherpe inzichten in de menselijke natuur, zou waarschijnlijk diepere lagen van de Vlaamse cultuur en identiteit verkennen, terwijl Jan Jambon, met zijn politieke achtergrond, wellicht een discussie zou voeren over de huidige stand van zaken en de toekomst van Vlaanderen.} \\  
        \midrule
        Llama-3-8B-dutch (ReBatch) & \textit{Het is niet mogelijk om met zekerheid te zeggen wat Jacques Brel, Willem Elsschot en Jan Jambon zouden bespreken als ze samen aan het café zouden zitten. Jacques Brel en Willem Elsschot zijn beiden overleden en Jan Jambon is een huidig politicus. Hun interesses en onderwerpen van gesprekken zijn afhankelijk van hun persoonlijke voorkeuren, de tijd die ze samen doorbrengen, en de context van hun gesprek. Echter, als we speculeren, kunnen we aannemen dat hun gesprekken zouden kunnen richten op onderwerpen zoals Vlaamse cultuur, literatuur, politiek, muziek, en misschien zelfs over hun eigen ervaringen en carrières. In het bijzonder zouden ze kunnen praten over hun respectievelijke bijdragen aan de Vlaamse cultuur en hoe die elkaar hebben beïnvloed of kunnen versterken.} \\
        \midrule
         \bottomrule
    \end{tabular}
    \label{tab:examples-other}
\end{table}

\end{document}